\documentclass[11pt]{article}

    \usepackage[utf8]{inputenc}
    \usepackage[T1]{fontenc}
    \usepackage[english]{babel}
    
    \usepackage{amsmath, amssymb, amsthm}
    \usepackage{graphicx}
    \usepackage{booktabs}
    \usepackage{multirow}
    \usepackage{hyperref}
    \usepackage{url}
    
    \usepackage[a4paper,margin=1in]{geometry}
    
    \usepackage{caption}
    \usepackage{subcaption}
    
    \title{LLM Modules:\\ Knowledge Transfer from a Large to a Small Model using Enhanced Cross-Attention}
    \author{Konstantin Kolomeitsev\\
    Almaty, Kazakhstan\\
    \texttt{uol92kot@gmail.com}}
    \date{\today}
    
\begin{document}
    
    \maketitle
    
    \begin{abstract}
    In this paper, we propose an architecture of \textit{LLM Modules} that enables the transfer of knowledge from a pre-trained large model to a smaller model using an Enhanced Cross-Attention mechanism. In the proposed scheme, the Qwen2-1.5B model is frozen, and its representations are passed through specially designed attention layers to the GPT-Neo-125M model, which is trained on limited computational resources. Experimental results obtained on the \texttt{Bespoke-Stratos-17k} dataset demonstrate that after 15 epochs of training, the combined model generates responses that are comparable in quality to models obtained by distillation. The paper discusses the advantages of the modular approach in detail, provides examples of input queries and their comparative analysis, and outlines prospects for further extension of the method.
    \end{abstract}
    
    \section{Introduction}
    Large language models (LLMs) have demonstrated outstanding performance in natural language processing tasks; however, their training and deployment require significant computational resources. This has led to the need for methods that transfer knowledge from large pre-trained models to smaller models. Such approaches are especially relevant for applied tasks with limited computational resources.
    
    In this work, we propose a modular LLM architecture in which a large model serves as a knowledge source, while a smaller model receives external representations via Enhanced Cross-Attention and generates responses. This method significantly reduces training costs while remaining effective for solving specific business tasks.
    
    \section{Related Work}
    Knowledge distillation and adaptive transfer of representations are widely discussed in the literature \cite{brown2020language, vaswani2017attention}. For instance, distillation works such as DistilBERT \cite{sanh2019distilbert} and TinyBERT \cite{jiao2019tinybert} show that it is possible to significantly reduce model size without losing performance. Modern transfer learning approaches include the use of adapter modules \cite{houlsby2019parameter} and methods like LoRA \cite{hu2021lora}, which allow for the adaptation of large models with a minimal number of trainable parameters. AdapterFusion \cite{pfeiffer2020adapterfusion} demonstrates the potential for non-destructive task composition during knowledge transfer. Additionally, the T5 work \cite{raffel2019exploring} explores the limits of transfer learning by unifying multiple tasks into a single text-to-text format. Thus, the proposed methodology differs in that, rather than relying on classic distillation, it transfers representations via specialized Cross-Attention layers, thereby preserving a larger portion of the original model's information and handling longer input sequences (e.g., 128K tokens in Qwen2 versus 2K tokens in GPT-Neo).
    
    \section{Methodology}
    \subsection{Core Idea of LLM Modules}
    The approach is based on two key components:
    \begin{itemize}
        \item \textbf{Knowledge Source:} A large pre-trained model (e.g., Qwen2-1.5B) is used to extract rich representations of the input query. The model's weights remain unchanged (frozen).
        \item \textbf{Generation Module:} A small model (e.g., GPT-Neo-125M) receives external representations via Enhanced Cross-Attention layers and, combining them with its own computations, generates a coherent response.
    \end{itemize}
    
    \subsection{Enhanced Cross-Attention}
    The key component of the scheme is a modified Cross-Attention layer, which consists of:
    \begin{enumerate}
        \item Linear projections that convert the representation dimensions of the large model (e.g., 1536) to the dimensions required by the small model (e.g., 768).
        \item An \textbf{Adapter Block} that provides an additional non-linear transformation to adapt the representations.
        \item A \textbf{Gating Mechanism} that dynamically blends the original representations with the external knowledge.
    \end{enumerate}
    This approach effectively "absorbs" knowledge from the source without the need to retrain the entire model.
    
    \section{Implementation}
    \subsection{Model Architecture}
    The architecture comprises three interconnected modules:
    \begin{enumerate}
        \item \textbf{ModifiedQwenWithCrossAttention:} 
        \begin{itemize}
            \item Loads the pre-trained Qwen2-1.5B model.
            \item Freezes all model parameters to preserve the pre-trained knowledge.
            \item Adds linear projections and Enhanced Cross-Attention layers to transform the hidden representations.
        \end{itemize}
        \item \textbf{ModifiedGptNeo:} 
        \begin{itemize}
            \item Based on the GPT-Neo-125M model.
            \item Freezes the embedding layers and replaces the final linear layer to align with the Qwen2 tokenizer.
        \end{itemize}
        \item \textbf{CombinedModel:} 
        \begin{itemize}
            \item Integrates the previous two modules.
            \item Receives an input query that is tokenized using the Qwen2 tokenizer.
            \item Passes the representations through the ModifiedQwenWithCrossAttention module, which are then processed by the ModifiedGptNeo model to generate the final response.
        \end{itemize}
    \end{enumerate}
    
    \subsection{Training and Data Preparation Details}
    The training dataset used is \texttt{Bespoke-Stratos-17k}, with a pre-filtering step to limit the maximum sequence length to 4096 tokens. The data preparation procedure includes:
    \begin{itemize}
        \item Dynamic padding of sequences using a \texttt{collate\_fn} function.
        \item Filtering out examples exceeding the specified length to avoid memory overflow errors.
        \item Splitting the dataset into training and validation sets using a random shuffling method.
    \end{itemize}
    Training is performed using the AdamW optimizer, and early stopping is applied to prevent overfitting.
    
    \subsection{Code Example}
    Below is a simplified code example for creating the combined model:
    \begin{verbatim}
    % Initialize configurations
    qwen_config = AutoConfig.from_pretrained("Qwen/Qwen2-1.5B")
    gptneo_config = AutoConfig.from_pretrained("EleutherAI/gpt-neo-125M")
    
    % Load the tokenizer
    qwen_tokenizer = AutoTokenizer.from_pretrained("Qwen/Qwen2-1.5B")
    if qwen_tokenizer.pad_token is None:
        qwen_tokenizer.pad_token = qwen_tokenizer.eos_token
    
    % Create the combined model
    model = CombinedModel(qwen_config, gptneo_config, qwen_tokenizer)
    
    % Initialize the optimizer to update only the trainable parameters
    optimizer = AdamW([
        {'params': model.qwen.cross_attention_layers.parameters(), 'lr': 1e-4},
        {'params': model.qwen.intermediate_layers.parameters(), 'lr': 1e-4},
        {'params': model.gptneo.parameters(), 'lr': 5e-5}
    ])
    \end{verbatim}
    
    \section{Experimental Study}
    \subsection{Comparative Analysis of Models}
    To evaluate the effectiveness of the proposed approach, the following models were compared:
    \begin{enumerate}
        \item \textbf{DeepSeek R1 671B} --- a large model with high generation quality.
        \item \textbf{DeepSeek-R1-Distill-Qwen-1.5B-GGUF 32FP} --- a distilled model.
        \item \textbf{Qwen2-1.5B-Instruct-GGUF FP16} --- the knowledge source model.
        \item \textbf{GPT-Neo-125M} --- the original small model.
        \item \textbf{GPT-Neo-125M-fine-tuned} --- GPT-Neo-125M further trained (with a 2048-token limit).
        \item \textbf{GPT-Neo-125M-clean} --- a model trained from scratch.
        \item \textbf{CombinedModel} --- the proposed paired model.
    \end{enumerate}
    
    \subsection{Examples of Input Queries and Results}
    To assess generation quality, the following test examples were used:
    \begin{enumerate}
        \item \textbf{Arithmetic Task:} \texttt{sum of 5 and 5}.\\
        All small models produced a set of incoherent text. Large pre-trained models generated the correct answer; however, the Qwen2 model did not provide any "reasoning" even when instructed. The CombinedModel demonstrated detailed reasoning comparable to large "reasoning" models.
        \item \textbf{Arithmetic Task:} \texttt{find the remainder by dividing 7 by 4}.\\
        The CombinedModel generated a response with a step-by-step explanation of the calculation, indicating the presence of a "reasoning" component.
    \end{enumerate}
    
    \subsection{Detailed Experimental Analysis}
    During the experiments, the following observations were made:
    \begin{itemize}
        \item After 15 epochs of training, the training loss and validation loss decreased significantly, from 13.8 to 2.3 in the first epoch and to 1.1 in subsequent epochs, confirming the successful convergence of the model.
        \item Despite the limited amount of training data, the CombinedModel demonstrated a significant improvement in quality compared to the original small models.
        \item Analysis of the generated responses (see Table~\ref{tab:results}) indicates that transferring knowledge via Enhanced Cross-Attention enables the combined model to produce more structured and logically coherent outputs.
    \end{itemize}
    
    \begin{table}[ht]
    \centering
    \resizebox{\textwidth}{!}{
    \begin{tabular}{l p{8cm} p{4cm}}
    \toprule
    Model & Response & Note \\
    \midrule
    DeepSeek R1 671B & [Detailed reasoning]\newline [Brief structured answer] & High computational complexity \\
    DeepSeek R1 Distill Qwen 1.5B & [Detailed reasoning]\newline [Brief structured answer] & Pre-trained model obtained via distillation \\
    Qwen2 1.5B & [Brief structured answer, without reasoning] & Pre-trained model \\
    GPT-Neo-125M-clean & [Lack of reasoning and coherent answer] & Trained from scratch \\
    GPT-Neo-125M-fine-tuned & [Lack of reasoning and coherent answer] & Improved via fine-tuning \\
    CombinedModel & [Detailed reasoning]\newline [Brief structured answer] & Optimized through knowledge transfer \\
    \bottomrule
    \end{tabular}
    }
    \caption{Comparison of Responses from Various Models}
    \label{tab:results}
    \end{table}
    
    \section{Discussion and Future Work}
    The advantages of the proposed approach include:
    \begin{itemize}
        \item A significant reduction in computational costs compared to full fine-tuning of large models.
        \item The ability to adapt the model to specific tasks by training it on only the necessary subset of data.
        \item The potential for exponential improvement in generation quality through the combination of representations from various architectures.
    \end{itemize}
    
    Future research directions include:
    \begin{itemize}
        \item Expanding the architecture by integrating other types of models (e.g., a combination of CNN and LLM).
        \item Investigating the impact of different configurations of Cross-Attention layers on generation quality.
        \item Applying the method to specialized business tasks to enhance adaptability and control over the generated responses.
    \end{itemize}
    
    \section{Conclusion}
    This work presents a novel approach to designing language models based on the use of \textit{LLM Modules} and an Enhanced Cross-Attention mechanism for transferring knowledge from a large pre-trained model to a smaller model. Experimental results demonstrate that even with limited computational resources, significant improvements in generation quality can be achieved, as evidenced by the reduction in training and validation loss and enhanced coherence of the generated responses. Further research in combining various architectures promises new opportunities for adapting models to specific tasks.
    
    \section*{Code and Data Availability}
    The source code, trained weights, and full outputs are available at:
    \url{https://huggingface.co/kkolomeitsev/llm-modules} (DOI: 10.57967/hf/4462).

    \end{document}